\pdfoutput=1
\documentclass[11pt]{article}

\usepackage{acl2022}

\usepackage{times}
\usepackage{latexsym}

\usepackage[T1]{fontenc}
\usepackage[utf8]{inputenc}

\usepackage{microtype}

\usepackage{enumitem}
\usepackage{amsmath}
\usepackage{amsfonts}
\usepackage{diagbox}
\usepackage{multirow}
\usepackage{threeparttable}
\usepackage{subcaption}
\usepackage{booktabs}       % professional-quality tables
\usepackage{nicefrac}       % compact symbols for 1/2, etc.
\usepackage{microtype}      % microtypography
\usepackage{algorithm}
\usepackage{algpseudocode}
\usepackage{xcolor}
\usepackage{graphicx}

\usepackage{makecell}

\title{Neural Label Search for Zero-Shot Multi-Lingual \\ Extractive Summarization}

\author{
  Ruipeng Jia$^{1,2}$\thanks{\; Work done during the first author’s internship at Microsoft Research Asia.},
  Xingxing Zhang$^{3}$\thanks{\; Corresponding authors},
  Yanan Cao$^{1,2}$\footnotemark[2],
  Shi Wang$^{4}$,
  Zheng Lin$^{1,2}$
  \textnormal{and} Furu Wei$^{3}$ \\
  $^{1}$Institute of Information Engineering, Chinese Academy of Sciences \\
  $^{2}$School of Cyber Security, University of Chinese Academy of Sciences \\
  $^{3}$Microsoft Research Asia \\
  $^{4}$Institute of Computing Technology, Chinese Academy of Sciences \\
  $^{1,2}$\texttt{\{jiaruipeng,caoyanan,linzheng\}@iie.ac.cn} \\
  $^{3}$\texttt{\{xizhang,fuwei\}@microsoft.com}, $^{4}$\texttt{wangshi@ict.ac.cn}
}

\begin{document}
\maketitle

\begin{abstract}
  In zero-shot multilingual extractive text summarization, a model is
  typically trained on English summarization dataset and then applied on
  summarization datasets of other languages.
  Given English gold summaries and documents, sentence-level labels for
  extractive summarization are usually generated using heuristics.
  However, these monolingual labels created on English datasets may not be
  optimal on datasets of other languages, for that there is the syntactic or
  semantic discrepancy between different languages.
  In this way, it is possible to translate the English dataset to other
  languages and obtain different sets of labels again using heuristics.
  To fully leverage the information of these different sets of labels, we
  propose NLSSum (\textbf{N}eural \textbf{L}abel \textbf{S}earch for
  \textbf{Sum}marization), which jointly learns hierarchical weights for these
  different sets of labels together with our summarization model.
  We conduct multilingual zero-shot summarization experiments on MLSUM and
  WikiLingua datasets, and we achieve state-of-the-art results using both
  human and automatic evaluations across these two datasets.
\end{abstract}

\section{Introduction}

The zero-shot multilingual tasks, which aim to transfer models learned on a
high-resource language (e.g., English) to a relatively low-resource language
(e.g., Turkish) without further training, are challenging
\cite{ruder2019survey}.
Recently, large pre-trained multilingual transformers such as M-BERT
\cite{devlin2018bert}, XLM \cite{lample2019cross}, and XLM-R
\cite{conneau2020unsupervised} have shown remarkable performance on zero-shot
multilingual natural language understanding tasks.
During pre-training, these transformer models project representations of
different languages into the same vector space, which makes the transfer learning
across different languages easier during fine-tuning \cite{gong2021lawdr}.
In zero-shot extractive summarization, we train an extractive model (based on a
pre-trained multilingual transformer) on English summarization dataset, which
selects important sentences in English documents.
Then, we apply this trained model to documents of a different language (i.e.,
extracting sentences of documents in another language).
In this paper, we aim to enhance the zero-shot capabilities of multilingual
sentence-level extractive summarization.

\begin{table}
  \scriptsize
  \begin{center}
    \begin{tabular}{|p{7cm}|}
      \hline
      \textbf{Sentence (English, Label 1):}
      \textit{\color{red} He was never charged in} that Caribbean Nation. \\
      \textbf{Reference Summary (English):}
      \textit{\color{red} He was} arrested twice, but \textit{\color{red} never charged in} Natalee Holloway's disappearance. \\
      \hline
      \textbf{Translated Sentence (German, Label 0):}
      \textit{\color{red} Er} wurde jedoch \textit{\color{red} nie} in dieser karibischen Nation \textit{\color{red} angeklagt}. \\
      \textbf{Translated Reference Summary (German):}
      Beim Verschwinden von Natalee Holloway wurde \textit{\color{red} er} zweimal verhaftet, aber \textit{\color{red} nie angeklagt}. \\
      \hline
    \end{tabular}
  \end{center}
  \caption{Monolingual Bias for Different Languages.}
  \label{example}
\end{table}

In text summarization, most datasets only contain human-written
\emph{abstractive} summaries as ground truth.
We need to transform these datasets into \emph{extractive} ones.
Thus, a greedy heuristic algorithm \cite{nallapati2017summarunner} is employed
to add one sentence at a time to the candidate extracted summary set, by
maximizing the ROUGE \cite{lin2004rouge} between candidate summary set and the
gold summary.
This process stops when none of the remaining sentences in the document can
increase the ROUGE anymore.
These selected sentences are labelled as one and all the other sentences labeled
as zero.
While the labels obtained from this greedy algorithm are monolingual-oriented and
may not be suitable for multilingual transfer.
For the example in Table \ref{example}, the English sentence is quite likely to
be selected as a summary sentence, since it greatly overlaps with the English
reference (high ROUGE).
While when the document and the summary are translated into German, the ROUGE
between the sentence and the summary is significantly lower (fewer $n$-gram
overlap).
Then, another sentence will be selected as substitution.
The greedy algorithm yields different labels on the English data and the
translated data and these labels may complement for each other.
We define this discrepancy as \emph{monolingual label bias}, and it is the key
to further improve the performance of zero-shot multilingual summarization.

To address the above problem, we design a method to create multiple sets of
labels with different machine translation methods according to the English
summarization dataset, and we employ NLSSum (\textbf{N}eural \textbf{L}abel
\textbf{S}earch for \textbf{Sum}marization) to search suitable weights for these
labels in different sets.
Specically, in NLSSum, we try to search the hierarchical weights (sentence-level
and set-level) for these labels with two neural weight predictors and these
label weights are used to train our summarization model.
During training, the two neural weight predictors are jointly trained with the
summarization model.
NLSSum is used only during training and during inference, we simply
apply the trained summarization model to documents in another language.

Experimental results demonstrate the effectiveness of NLSSum, which
significantly outperforms original XLMR by 2.25 ROUGE-L score on MLSUM
\cite{scialom2020mlsum}.
The human evaluation also shows that our model is better  compared to other
models.
To sum up, our contributions in this work are as follows:

\begin{itemize}
\item To the best of our knowledge, it is the first work that studies the
  \emph{monolingual label bias} problem in zero-shot multilingual extractive
  summarization.

\item We introduce the multilingual label generation algorithm (Section
  \ref{sec_labels}) to improve the performance of multilingual zero-shot models.
  Meanwhile, we propose the NLSSum architecture (Section \ref{sec_sent_level})
  to search suitable weights for different label sets.

\item Extensive experiments are conducted with detailed analysis, and the
  results across different datasets demonstrate the superior performance on
  multilingual datasets.
  In MLSUM, the zero-shot performance on Russian is even close to its
  supervised counterpart.

\end{itemize}

\section{Related Work}

There has been a surge of research on multilingual pretrained models, such as
multilingual BERT \cite{devlin2018bert}, XLM \cite{lample2019cross} and
XLM-RoBERTa \cite{conneau2020unsupervised}.
For multilingual summarization, the summarize-then-translate and
translate-then-summarize are widely used approaches in prior studies
\citet{lim2004multi}.
There is another effective multi-lingual data augmentation, a method that
replaces a segment of the input text with its translation in another language
\cite{singh2019xlda}.

On the other hand, large-scale multilingual summarization datasets have been
introduced \cite{scialom2020mlsum,ladhak2020wikilingua}, which enable new
research directions for the multilingual summarization.
\citet{nikolov2020abstractive} applies an alignment approach to collect
large-scale parallel resources for low-resource domains and languages.
In this paper, we aim to advance the multilingual zero-shot transferability, by
training extractive summarization on English and inferring on other languages.

\section{Methodology}
\begin{figure}
  \centering
  \includegraphics[height=5.6cm]{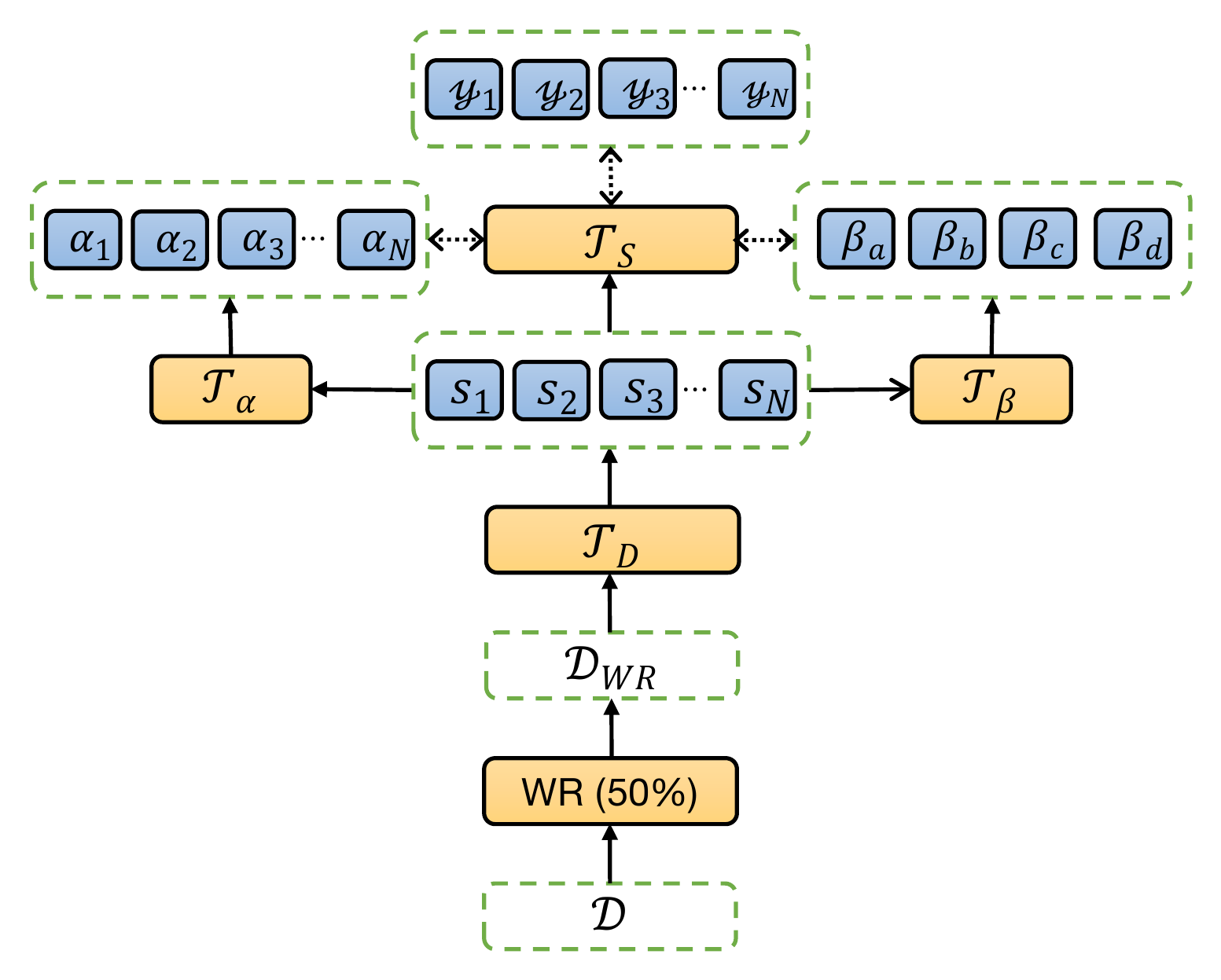}
  \caption{Overview of NLSSum. The input English document is argumented by 50\%
    word replacement and the output is supervised by multilingual labels.}
  \label{figure_architecture}
\end{figure}

\subsection{Problem Definition}
\label{sec_problem}

Let $\mathcal{D} = (s_1, s_2, ..., s_N)$ denotes a document with $N$ sentences,
where $s_i = (w_{1}^i, w_{2}^i, ..., w_{|s_i|}^i)$ is a sentence in
$\mathcal{D}$ with \(|s_i|\) words.
$\mathcal{S}$ is the human-written summary.
Extractive summarization can be considered as a sequence labeling task that
assigns a label \(y_i \in \{0, 1\}\) to each sentence \(s_i\), where \(y_i = 1\)
indicates the \(i\)-th sentence should be included in the extracted summary.
The gold labels of sentences in $\mathcal{D}$ are obtained from
(\textit{\(\mathcal{D}\), \(\mathcal{S}\)}) by the greedy heuristic algorithm
\cite{nallapati2017summarunner}, which adds one sentence at a time to the
extracted summary, skipping some sentences to maximize the ROUGE score of
\(\mathcal{S}\) and the extracted sentences.
In multi-lingual zero-shot setting, the summarization model is trained on
English dataset and is finally applied on documents of other languages.

\subsection{Neural Extractive Summarizer}
\label{sent_encoder}
Our sentence encoder builds upon the recently proposed XLMR
\cite{conneau2020unsupervised} architecture, which is based on the deep
bidirectional Transformer \cite{vaswani2017attention} and has achieved
state-of-the-art performance in many multilingual zero-shot understanding tasks.
Our extractive model is composed of a sentence-level Transformer
\(\mathcal{T}_S\) (initialized with XLMR) and a document-level Transformer
\(\mathcal{T}_D\) (a two-layer Transformer).

For each sentence \(s_i\) in the input document \(\mathcal{D}\),
\(\mathcal{T}_S\) is applied to obtain a contextual representation for each word
\(w_{j}^i\):

\begin{equation}
  [\mathbf{u}_{1}^1, \mathbf{u}_{2}^1, ..., \mathbf{u}_{|s_N|}^N] = \mathcal{T}_S([w_{1}^1, w_{2}^1, ..., w_{|s_N|}^N])
  \label{eq_word}
\end{equation}

Similar to \newcite{liu2019text}, the representation of a sentence \(s_i\)
is acquired by taking the representation of the first token in the sentence \(\mathbf{u}_{1}^i\).
The document-level Transformer \(\mathcal{T}_D\) (a two-layer inter-sentence
Transformer), which is stacked to \(\mathcal{T}_S\), takes \(s_i\) as input and yields a contextual
representation \(\mathbf{v}_i\) for each sentence.
We intend this process to further captures the sentence-level features for extractive summarization:
\begin{equation}
  [\mathbf{v}_1, \mathbf{v}_2, ..., \mathbf{v}_N] = \mathcal{T}_D ([\mathbf{u}_{1}^1, \mathbf{u}_{1}^2, ..., \mathbf{u}_{1}^N])
  \label{eq_sent}
\end{equation}

\begin{figure}
  \centering
  \includegraphics[height=4.4cm]{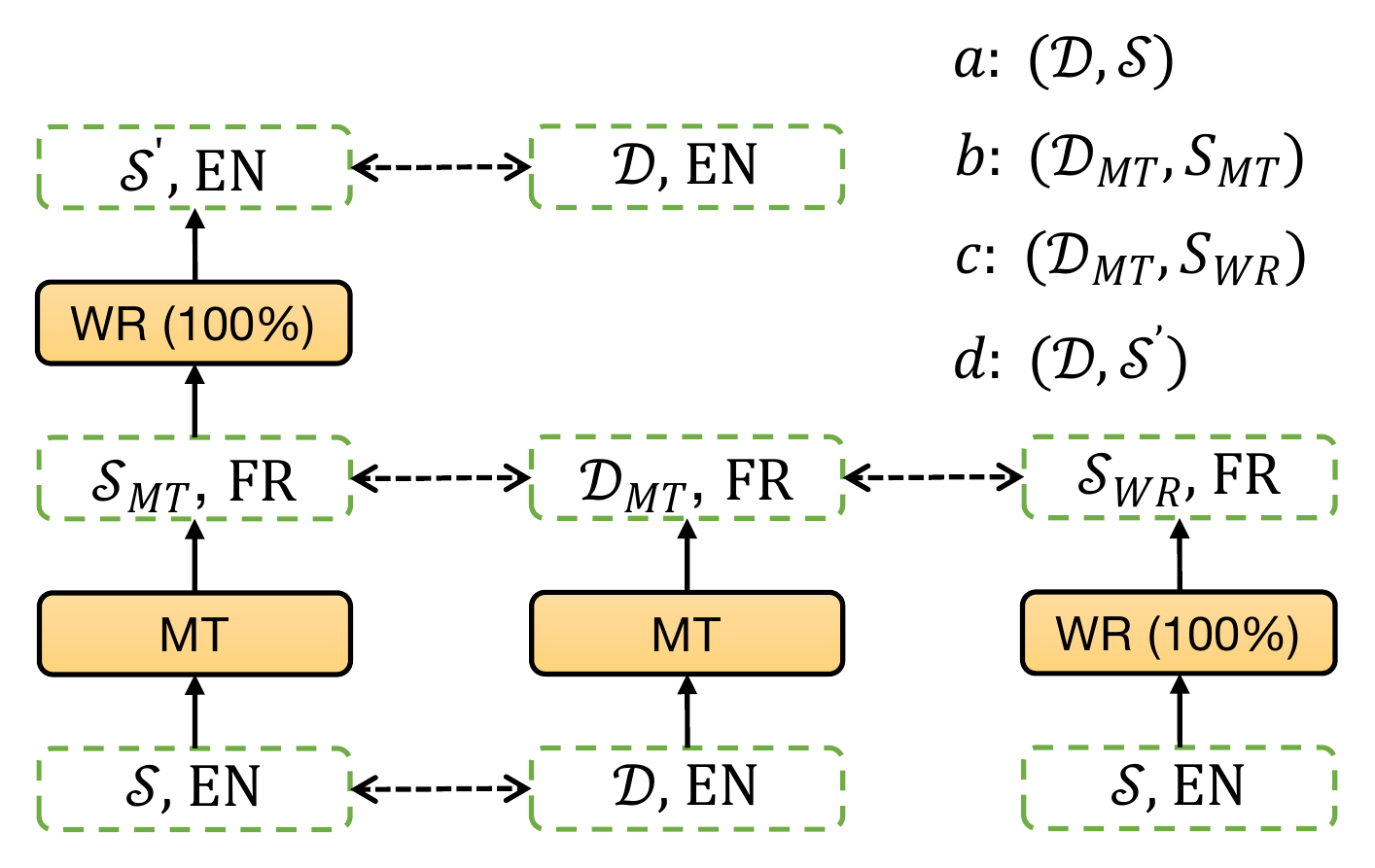}
  \caption{Four Sets of Multilingual Label.
    `EN' is the symbol of English and `FR' is for the foreign language.}
  \label{figure_label}
\end{figure}

For sentence \(s_i\), the final output prediction of the extractive model
$\hat{y}_i$ (i.e., the probability of being selected as summary) is obtained
through a linear and a sigmoid classifier layer:
\begin{equation}
  \label{eq_pred}
  \hat{y}_i = \sigma(\mathbf{W}_o \mathbf{v}_i + b_o)
\end{equation}
where \(\mathbf{W}_o\) and \(b_o\) are the weight matrix and bias term. Next we
introduce how we obtain the neural labels for model training.

\subsection{Overview of Neural Label Search}
The training and inference of our NLSSum model includes five steps as follows.

\begin{enumerate}[label=(\Roman*),leftmargin=2\parindent]
  \item \textbf{Multilingual Data Augmentation}:
  This step aims to enhance the multilingual transfer capability of our
  extractive model and alleviate the discrepancy between training (on English)
  and inference (on unseen languages).

  \item \textbf{Multilingual Label Generation}:
    The extractive model is supervised by multilingual label, which consists of
    four sets of labels, according to different strategies.

  \item \textbf{Neural Label Search}:
    In this step, we design the hierarchical sentence-level and set-level
    weights for labels of different strategies.
    The final weights are calculated with a weighted average and assigned to
    corresponding sentences.

  \item \textbf{Fine-Tuning}:
    We fine-tune our extractive model the augmented English document (generated
    in Step I) with supervision from the weighted multilingual labels (generated
    in Step III), as shown in Figure \ref{figure_architecture}.

  \item \textbf{Zero-Shot}:
    We apply the model fine-tuned on English data (Step IV) to extract sentences
    on documents of the target language.
\end{enumerate}

\subsection{Multilingual Data Augmentation}
\label{sec_data_aug}
In the training process, only the raw English documents and its paired summary
labels are available.
We use the following two methods for multilingual data argumentation of English
documents, which we intend the model to align its English representations with
representations in other languages.

\paragraph{Word Replacement (WR)}
Similar to \citet{qin2020cosdaml}, we enhance multilingual transferability by
constructing \textit{Word Replacement} data in multiple languages
\emph{dynamically}.
Let FR denote a foreign language.
Specifically, a set of words are randomly chosen in raw English documents and
replaced with words in FR using the bilingual dictionary MUSE
\cite{conneau2018word}.
This approach can in some degree align the replaced word representations in FR
with their English counterpart by mixing with the English context.

\paragraph{Machine Translation (MT)}
The above augmentation method is applied \emph{dynamically} during training, and
\textit{Machine Translation} yet is another offline strategy to augment data.
First, we translate documents and their paired summaries from English into the
target language FR using the MarianMT
system\footnote{https://github.com/marian-nmt/marian}
\cite{junczysdowmunt2018marian}.
Then, the labels are generated on the translated data with the same greedy
algorithm as on English data.
Finally, the extractive model is fine-tuned on the translated documents with the
supervision of new labels, and inferred on the original FR document.

Unfortunately, the performance of machine translation is instable with the noise
or error propagation \cite{wan2010cross}.
Therefore, we choose the word replacement method here to enhance the input
document and the argumented document is served as the input of our extractive
model.
Note that we do use both the word replacement and machine translation methods to
generate multilingual labels (see the next section).

\subsection{Multilingual Labels}
\label{sec_labels}

Given an English article $\mathcal{D}$ and its summary $\mathcal{S}$, we can
obtain its extractive labels using the greedy algorithm introduced in Section
\ref{sec_problem}.

\paragraph{Label Set $U_a$} Let \mbox{ $U_a = {\tt GetPosLabel}(\mathcal{D},
  \mathcal{S}) $ } denote the indices of sentences with positive labels, where
${\tt GetPosLabel}(\mathcal{D}, \mathcal{S})$ returns the indices of positive
labeled sentences in the original English document $\mathcal{D}$ using the
greedy algorithm.
The labels created on English data $(\mathcal{D}, \mathcal{S})$ may not be
optimal in multilingual settings (inference on a different language).
As shown in Figure \ref{figure_label}, we therefore create yet another three
label sets using the WR and MT methods introduced earlier to simulate the
multilingual scenario during inference time.

\paragraph{Label Set $U_b$} To create labels based foreign language (FR) data,
we translate both the English document $\mathcal{D}$ and its summary
$\mathcal{S}$ to FR using the MT method in Section \ref{sec_data_aug}, resulting
$\mathcal{D}_{MT}$ and $\mathcal{S}_{MT}$ (also see Figure \ref{figure_label}).
Again by using the greedy algorithm, we obtain the indices of sentences with
positive labels $U_b = {\tt GetPosLabel}(\mathcal{D}_{MT}, \mathcal{S}_{MT})$.

\paragraph{Label Set $U_c$} Label set $U_c$ is also based on FR data.
To make label set $U_c$ different from $U_b$, we translate $\mathcal{D}$ to
$\mathcal{D}_{MT}$ using the MT method, while we translate $\mathcal{S}$ to
$\mathcal{S}_{WR}$ using the WR method (we do 100\% word replacement) with the
EN-FR dictionary.
The resulting label set $U_c = {\tt GetPosLabel}(\mathcal{D}_{MT}, \mathcal{S}_{WR})$.

\paragraph{Label Set $U_d$} Label set $U_d$ is based on English data.
The idea is to create a paraphrased English summary $\mathcal{S}'$ using the
back translation technology.
We first translate $\mathcal{S}$ to $\mathcal{S}_{MT}$ using MT method and
translate $\mathcal{S}_{MT}$ back to English $\mathcal{S}'$ using the WR method
(100\% word replacement).
We use different \emph{translation} method for forward and backward translations
to maximize the different between $\mathcal{S}$ and $\mathcal{S}'$.
Finally, $U_d = {\tt GetPosLabel}(\mathcal{D}, \mathcal{S}')$.

Note that there are also many other possible strategies for creating
multilingual labels and we only use these four strategies above as examples to
study the potential of multilingual labels.
Intuitively, the contributions of these four label sets for multilingual
transferability are different, and the MT and WR translation methods may
introduce translation errors, which result noisy labels.
Therefore, we introduce the \textit{Neural Label Search} in the next section to
find suitable weights for these multilingual labels.

\subsection{Neural Label Search}
\label{sec_sent_level}

In this section, we assign a weight for each sentence in a document and the
weight will be used as the supervision to train our extractive model.
Note that the weight is a multiplication of a sentence level weight and a label
set level weight.
Let \(\mathcal{T}_{\alpha}\) denote the sentence level weight predictor and
\(\mathcal{T}_{\beta}\) the set level weight predictor.
The implementation of \(\mathcal{T}_{\alpha}(\cdot) = \sigma( g(
\mathcal{T'}_{\alpha}(\cdot) ) ) \) is a two-layer transformer model
$\mathcal{T'}_{\alpha}(\cdot)$ followed by a linear layer $g(\cdot)$ and a
sigmoid function.
The implementation of \(\mathcal{T}_{\beta}\) is the same as
\(\mathcal{T}_{\alpha}\), but with different parameters.

The predictor \(\mathcal{T}_{\alpha}\) transforms sentence representations (see
Equation (\ref{eq_word}) for obtaining $\mathbf{u}_j^i$) to probabilities
\(\alpha_i \in [0, 1]\) as follows:
\begin{equation}
  \begin{split}
 & [\hat{\alpha_1}, \hat{\alpha_2}, ..., \hat{\alpha_N}] = \mathcal{T}_{\alpha}([\mathbf{u}_1^1, \mathbf{u}_1^2, ..., \mathbf{u}_1^N]) \\
  & \alpha_i =
    \begin{cases}
      \hat{\alpha_i}, & \text{if } i \in U \\
      0,        & \text{otherwise}
    \end{cases}
  \end{split}
  \label{eq_alpha}
\end{equation}
where $U = U_a \cup U_b \cup U_c \cup U_d$.
Note that we only predict weights for sentences with non-zero labels, since we
believe that these sentences, which are the minority, are more informative than
zero-label sentences.

The computation of \(\mathcal{T}_{\beta}\) is similar, but we first do a mean
pooling over sentences in each label set.

\small
\begin{equation*}
  [\beta_a, \beta_b, \beta_c, \beta_d] = \mathcal{T}_{\beta}([\frac{\sum\limits_{i \in U_a} \mathbf{u}_1^i}{n_a}, \frac{\sum\limits_{i \in U_b} \mathbf{u}_1^i}{n_b}, \frac{\sum\limits_{i \in U_c} \mathbf{u}_1^i}{n_c}, \frac{\sum\limits_{i \in U_d} \mathbf{u}_1^i}{n_d}])
  \label{eq_beta}
\end{equation*}
\normalsize
where \(n_a, n_b, n_c, n_d\) are sizes of the four label sets.

The final weight $l_i$ for sentence $s_i$ is 0 when $i \notin U$ ($i$ does not
belong to any label set). Otherwise, the computation of $l_i$ is as follows.
\begin{equation}
  l_i = \alpha_i * \frac{\sum_{j \in \{a,b,c,d\}} \beta_j^i}{m_i}
  \label{eq_beta}
\end{equation}
where if \(i \in U_j\), \(\beta_j^i\) is \(\beta_j\), else \(\beta_j^i\) is \(0\) and
\(m_i\) is the number of label sets containing $i$.
Note that one sentence may belong to multiple label sets, so we normalize its
$\beta_j^i$ weights in Equation (\ref{eq_beta}).

\paragraph{Weight Normalization}
In this paper, we only calculate the multilingual weights for multilingual
labels, in which the corresponding sentences are all selected as summary
sentences by different document-summary pairs, as shown in the Figure
\ref{figure_label}.
The label weights $l_i$ are used to train our summarization model, whose output
$\hat{y}_i$ is through a sigmoid function (Equation \ref{eq_pred}).
$\hat{y}_i > 0.5$ means sentence $s_i$ could be selected as in summary.
Therefore, when $i \in U$, we rescale $l_i$ to $[0.5, 1.0]$:

\begin{equation}
  \begin{split}
    l_i = \frac{l_i - l_{min}}{2 * (l_{max} - l_{min})} + 0.5
  \end{split}
  \label{eq_all}
\end{equation}
where \(l_{max}\) and \(l_{min}\) are the maximum and minimum value of \(l_i\), when \(i \in U\).

\subsection{Training and Zero-shot Inference}
In this section, we present how we train our extractive model as well as the two
weight predictors \(\mathcal{T}_{\alpha}\) and \(\mathcal{T}_{\beta}\).
Note that we train the components above jointly.
We train the extractive model using both the English labels $y^a$ (created using
the greedy algorithm) as well as the label weights generated in Section
\ref{sec_sent_level}.
To train \(\mathcal{T}_{\alpha}\), we use binary labels $y^{\alpha}$, where in
one document, $y_i^{\alpha} = 1$ when $i \in U$, otherwise $y_i^{\alpha} = 0$.
To train \(\mathcal{T}_{\beta}\), we again use binary labels $y^{\beta}$, but
these labels are on set level rather than sentence level.
Defining positive examples for \(\mathcal{T}_{\beta}\) is straight-forward and
we set $y^{\beta}_q = 1$ when $q \in \{U_a, U_b, U_c, U_d\}$ (each label set
corresponds to one positive example).
For negative examples in one particular document, we randomly sample three
sentence indices from sentences with zero labels as one negative example.
We finally make the numbers of positive and negative examples for
\(\mathcal{T}_{\beta}\) close to 1:1.

The final loss is a sum of the four losses above:
\begin{equation}
  \begin{split}
    \mathcal{L} = & CE(\hat{y}, y^a) + CE(\hat{y}, l) + \\
                  & CE(\alpha, y^{\alpha}) + CE(\beta, y^{\beta})
  \end{split}
\label{equation_ce}
\end{equation}
where $CE$ is the cross entropy loss; $l$ is the weighted multilingual label
(Section \ref{sec_sent_level}); \(y^a\), \(y^{\alpha}\), and
\(y^{\beta}\) are the binary labels for the supervision of \(\hat{y}\),
\(\alpha\), and \(\beta\).
Specifically, $\alpha = [\alpha_1, \alpha_2, \dots, \alpha_N]$ and $\beta =
[\beta_a, \beta_b, \beta_c, \beta_d]$ (just as Equation \ref{eq_alpha} and \ref{eq_beta}).

During the zero-shot inference, we simply apply the model trained on the English
dataset using the objectives above to other languages.

\begin{table}
  \begin{center}
    \resizebox{0.8\columnwidth}!{
    \begin{tabular}{l|c}
      \toprule
      \multirow{2}*{Datasets}  & \multirow{2}*{\# Docs (Train / Val / Test)}  \\
                               &                                              \\
      \midrule
      CNN/DM, English          &  287,227 / 13,368 / 11,490                    \\
      MLSUM, German            &  220,887 / 11,394 / 10,701                    \\
      MLSUM, Spanish           &  266,367 / 10,358 / 13,920                    \\
      MLSUM, French            &  392,876 / 16,059 / 15,828                    \\
      MLSUM, Russian           &  25,556 / 750 / 757                           \\
      MLSUM, Turkish           &  249,277 / 11,565 / 12,775                    \\
      \midrule
      WikiLingua, English      &  99,020 / 13,823 / 28,614                     \\
      WikiLingua, German       &  40,839 / 5,833 / 11,669                      \\
      WikiLingua, Spanish      &  79,212 / 11,316 / 22,632                     \\
      WikiLingua, French       &  44,556 / 6,364 / 12,731                      \\
      \bottomrule
    \end{tabular}
    }
  \end{center}
  \caption{Data Statistics: CNN/Daily Mail, MLSUM and WikiLingua.}
  \label{dataset}
\end{table}

\section{Experiments}

\subsection{Datasets}
\paragraph{MLSUM \& CNN/DM}
MLSUM is the first large-scale multilingual summarization dataset
\cite{scialom2020mlsum}, which is obtained from online newspapers and contains
1.5M+ document/summary pairs in five different languages, namely, French(Fr),
German(De), Spanish(Es), Russian(Ru), and Turkish(Tr). The English dataset is the popular CNN/Daily mail (CNN/DM) dataset \cite{hermann2015teaching}. Our model is trained on CNN/DM.

\paragraph{WikiLingua}
A large-scale, cross-lingual dataset for abstractive summarization
\cite{ladhak2020wikilingua}.
The dataset includes 770K article and summary pairs in 18 languages from
WikiHow\footnote{https://www.wikihow.com}.
Our training setting is identical to that of MLSUM, our extractive model is trained on the English
data and inferred on other three languages (French, German, Spanish).
MLSUM and WikiLingua are described in detail in Table \ref{dataset}.

\begin{table}[t]
  \begin{center}
    \resizebox{1.0\columnwidth}!{
    \begin{tabular}{lrrrrr|r}
      % \toprule
      \Xhline{4\arrayrulewidth}
      \multirow{2}*{Models } & \multicolumn{6}{c}{MLSUM} \\
                                      & De              & Es              & Fr              & Ru              & Tr              & avg    \\
      \midrule
      Oracle\(^\star\)                & 52.30           & 35.78           & 37.69           & 29.80           & 45.78           & 40.27  \\
      Lead-2\(^\star\)                & 33.09           & 13.70           & 19.69           & 5.94            & 28.90           & 20.26  \\
      \Xhline{4\arrayrulewidth}
      \multicolumn{7}{c}{\textbf{Supervised}} \\
      \hline
      Pointer-Generator               & 35.08           & 17.67           & 23.58           & 5.71            & 32.59           & 22.99  \\
      mBERTSum-Gen                    & 42.01           & 20.44           & 25.09           & 9.48            & 32.94           & 25.99  \\
      XLMRSum\(^\star\)               & 41.28           & 21.99           & 24.12           & 10.44           & 33.29           & 26.22  \\
      MARGE (Train One)               & 42.60           & 22.31           & 25.91           & 10.85           & 36.09           & 27.55  \\
      MARGE (Train All)               & 42.77           & 22.72           & 25.79           & 11.03           & 35.90           & 27.64  \\
      \Xhline{4\arrayrulewidth}
      \multicolumn{7}{c}{\textbf{Zero-Shot}} \\
      \hline
      MARGE                           & 30.01           & 17.81           & 19.39           & 8.67            & 29.39           & 21.05  \\
      mBERTSum\(^\star\)              & 17.36           & 17.27           & 19.64           & 8.37            & 19.30           & 16.39  \\
      XLMRSum\(^\star\)               & 32.05           & 19.49           & 22.20           & 8.70            & 27.64           & 22.02  \\
      XLMRSum-MT\(^\star\) w/ $U_a$   & 29.34           & 21.14           & 23.82           & 8.68            & 24.23           & 21.44  \\
      XLMRSum-MT\(^\star\) w/ $U_b$   & 29.70           & 21.18           & 23.62           & 9.37            & 24.27           & 21.63  \\
      XLMRSum-WR\(^\star\)            & 32.37           & 21.03           & 23.67           & 9.34            & 30.10           & 23.30  \\
      NLSSum-Sep\(^\star\)            & 34.21           & \textbf{21.24}  & \textbf{23.92}  & 10.09           & \textbf{31.68}  & 24.23  \\
      NLSSum\(^\star\)                & \textbf{34.95}  & 21.20           & 23.59           & \textbf{10.13}  & 31.49           & \textbf{24.27}  \\
      % \bottomrule
      \Xhline{4\arrayrulewidth}
    \end{tabular}
    }
  \end{center}
  \caption{ROUGE-L on MLSUM dataset. $^\star$ means extractive models, and others are abstractive models. }
  \label{result_mlsum}
\end{table}

\subsection{Evaluation}
Similar to \citet{liu2019text}, we also select the top three sentences as the
summary, with Trigram Blocking to reduce redundancy.
Following \newcite{scialom2020mlsum}, we report the F1 ROUGE-L score of NLSSum
with a full Python implemented ROUGE
metric\footnote{https://github.com/pltrdy/rouge}, which calculates the overlap
lexical units between extracted sentences and ground-truth.
Following \citet{lin2004rouge}, to assess the significance of the results, we
applied bootstrap resampling technique \cite{davison1997bootstrap} to estimate
95\% confidence intervals for every correlation computation.

\subsection{Implementation}
Our implementation is based on Pytorch \cite{paszke2019pytorch} and
transformers.
The pre-trained model employed in NLSSum is {\tt XLMR-Large}.  % mbert have no large version
We train NLSSum on one Tesla V100 GPU for 100,000 steps (2 days) with
a batch size of 4 and gradient accumulation every two steps.
Adam with $\beta_1 = 0.9, \beta_2 = 0.999$ is used as optimizer.
The learning rate is linearly increased from 0 to $1e-4$ in the first 2,500
steps (warming-up) and linearly decreased thereafter.
For the source document data augmentation, we use a 0.5 word replacement rate
with a bilingual dictionary \cite{conneau2018word}.

\subsection{Models in Comparison}

\textbf{Oracle} sentences are extracted by the greedy algorithm introduced in Section
\ref{sec_problem}.
\textbf{Lead-K} is a simple baseline to choose the first k sentences in a document
as its summary. We use $k=2$ on MLSUM and $k=3$ on WikiLingua, which lead to the best results.
\textbf{Pointer-Generator} augments the standard Seq2Seq model with copy and
coverage mechanisms \cite{See2017PointerGenerator}.
\mbox{\textbf{mBERTSum-Gen}} is based on the multi-lingual version BERT (mBERT; \citealt{devlin2018bert}) and it is extended to do generation with a unified masking method in UniLM \cite{dong2019unified}.
\textbf{MARGE} is a pre-trained seq2seq model learned with an unsupervised multilingual
paraphrasing objective \cite{lewis2020pretraining}. \textbf{mBERTSum},
\textbf{XLMRSum},
\textbf{XLMRSum-MT} and \textbf{XLMRSum-WR} are all extractive models described in Section \ref{sent_encoder} and their sentence encoders are either initialized from {\tt mBERT} or {\tt XLMR-Large}.
They are all trained on the Enlgish dataset.
XLMRSum-MT is trained on the English training data argumented with machine translation. While XLMRSum-WR is trained on the English training data argumented with bilingual dictionary word replacement.

\section{Result \& Analysis}

\begin{table}[t]
  \begin{center}
    \resizebox{0.7\columnwidth}!{
      \begin{tabular}{lrrr|r}
        \Xhline{4\arrayrulewidth}
        \multirow{2}*{Models } & \multicolumn{4}{c}{WikiLingua} \\
                               & De              & Es              & Fr              & avg    \\
        \midrule
        Oracle      & 30.81           & 36.52           & 34.64           & 33.99  \\
        Lead-3      & 16.32           & 19.78           & 18.40           & 18.17  \\
        \midrule
        mBERTSum       & 18.83           & 22.49           & 20.91           & 20.74  \\
        XLMRSum     & 22.10           & 26.73           & 25.06           & 24.63  \\
        XLMRSum-MT  & 21.92           & 26.41           & 24.75           & 24.36  \\
        XLMRSum-WR  & 22.20           & 26.78           & 25.10           & 24.69  \\
        NLSSum   & \textbf{22.45}  & \textbf{26.98}  & \textbf{25.34}  & \textbf{24.92}  \\
        \Xhline{4\arrayrulewidth}
      \end{tabular}
    }
  \end{center}
  \caption{Zero-Shot ROUGE-L Results of WikiLingua}
  \label{result_wikilingua}
\end{table}

\paragraph{ROUGE Results on MLSUM}
Table \ref{result_mlsum} shows results on MLSUM.
The first block presents the Oracle upper bound and the Lead-2 baseline, while
the second block includes the supervised summarization results.
Results of Pointer-Generator, mBERTSum-Gen are reported in
\newcite{scialom2020mlsum}, while results of MARGE are reported in
\newcite{lewis2020pretraining}.
The results of MARGE training on all languages jointly (Train All) are slightly
better than its counterpart when training on each language separately (Train
One).
While we see a different trend with other models.
Comparing extractive models against abstractive models in the supervised
setting, the abstractive paradigm is still the better choice.

\begin{table}[t]
  \centering
  \resizebox{0.7\columnwidth}!{
    \begin{tabular}{lrrrrrr}
      \Xhline{4\arrayrulewidth}
      Models         & 1st  & 2nd  & 3rd  & 4th  & MeanR \\
      \midrule
      mBERTSum          & 0.07 & 0.25 & 0.31  & 0.37 & 2.98 \\
      XLMRSum        & 0.16 & 0.28 & 0.27  & 0.29 & 2.69 \\
      NLSSum         & 0.28 & 0.32 & 0.2  & 0.2 & 2.32 \\
      Oracle         & 0.49 & 0.15 & 0.22 & 0.14 & 2.01 \\
      \Xhline{4\arrayrulewidth}
    \end{tabular}
  }
  \caption{Human Evaluation on MLSUM, German}
  \label{human_evaluation}
\end{table}

We present the zero-shot results in the third block.
All models are trained on the Enlgish summarization dataset and infered on
dataset of other languages.
With a decent multi-lingual pre-trained model, the extractive XLMRSum performs
better than the abstractive MARGE, which demonstrates the superiority of
extractive approaches in zero-shot summarization.
When applying machine translation based (XLMRSum-MT) and multi-lingual word
replacement based (XLMRSum-WR) data argumentation method to XLMR (see Section
\ref{sec_data_aug}), we obtain further improvements.
With MT based argumentation method (XLMRSum-MT), we could re-generate extractive
labels using the translated doucments and summaries (the \(U_b\) setting).
We do observe that the re-generated labels could slightly improve the results,
but the resulting XLMRSum-MT is still worse than XLMRSum and XLMRSum-WR.
With the neural label search method, NLSSum-Sep outperforms all models in
comparison.
For faster feedback, we train a separate model for each language in XLMRSum-MT and
XLMRSum-WR and NLSSum-Sep (models for different languages can be trained in
parallel), which is to do data argumentation only to one target language.
In our final model NLSSum, we train one model for all languages (we do data
argumentation from English to all target languages) and we observe that the
results of NLSSum-Sep and NLSSum are similar.
Compared with the original XLMRSum, NLSSum achieves 2.27 improvements on the
average R-L score, which is a remarkable margin in  summarization.
It indicates that our multilingual neural label search method significantly
improves the multilingual zero-shot transferability.
The differences between NLSSum and other models in comparison except NLSSum-Sep
are significant (p < 0.05).
Specifically, the performance XLMRSum-MT is worse than that of XLMRSum.
For more in-depth analysis, we note that:
1) As the input of a model, the translation-based documents are prone to the
error propagation, therefore, we should avoid to encode these noise documents.
2) Fortunately, our multilingual label only applies the translation method when
converting document/summary pair into labels, instead of encoding.

\paragraph{ROUGE Results on WikiLingua}
To further evaluate the performance of NLSSum, we design
additional zero-shot experiments for all our extractive models on WikiLingua.
These models are trained on English and inferred on other three languages.
The results are in Table \ref{result_wikilingua}.
We observe that our NLSSum still performs better than all the other extractive models.
Meanwhile, compared with the results on MLSUM, the improvement on WikiLingua is
not remarkable.
Probably because the documents and summaries in WikiLingua are a series of
how-to steps, which are more platitudinous than news summarization.

\subsection{Ablation Studies}

\begin{table}[t]
  \begin{center}
    \resizebox{1.0\columnwidth}!{
    \begin{tabular}{lrrrrr|r}
      \Xhline{4\arrayrulewidth}
      \multirow{2}*{Models }                & \multicolumn{6}{c}{MLSUM} \\
                                            & De              & Es              & Fr              & Ru             & Tr              & avg    \\
      \midrule
      XLMRSum                               & 30.35           & 20.67           & 22.85           & 9.39           & 31.55           & 22.81  \\
      NLSSum \ w/o \(\mathcal{T}_{\beta}\)  & 33.13           & 21.21           & 23.09           & 9.72           & \textbf{32.68}  & 23.97  \\
      NLSSum                                & \textbf{33.51}  & \textbf{21.74}  & \textbf{24.10}  & \textbf{9.91}  & 32.58           & \textbf{24.37}  \\
      \Xhline{4\arrayrulewidth}
      \multicolumn{7}{c}{\textbf{Train with Different Label Sets}} \\
      \hline
      XLMRSum-WR w/ $U_a$                  & \textbf{32.09}  & \textbf{21.04}  & \textbf{23.33}  & 9.69           & \textbf{32.04}  & \textbf{23.58}  \\
      XLMRSum-WR w/ $U_b$                  & 30.39           & 20.71           & 23.17           & \textbf{9.83}  & 31.37           & 23.05  \\
      XLMRSum-WR w/ $U_c$                  & 29.66           & 20.64           & 22.96           & 9.32           & 31.63           & 22.76  \\
      XLMRSum-WR w/ $U_d$                  & 30.22           & 20.16           & 22.90           & 9.61           & 31.90           & 21.78  \\
      \Xhline{4\arrayrulewidth}
      \multicolumn{7}{c}{\textbf{Train with All Label Sets and with Fixed Weights}} \\
      \hline
      XLMRSum-WR, w=0.6                    & 32.12           & \textbf{21.05}  & 23.30           & 9.31           & 32.51           & 23.65  \\
      XLMRSum-WR, w=0.7                    & 32.46           & 20.73           & \textbf{23.67}  & \textbf{9.77}  & 32.72           & 23.82  \\
      XLMRSum-WR, w=0.8                    & \textbf{32.86}  & 20.98           & 23.42           & 9.64           & \textbf{32.93}  & \textbf{23.91}  \\
      XLMRSum-WR, w=0.9                    & 32.41           & 20.48           & 23.27           & 9.57           & 32.63           & 23.65  \\
      \Xhline{4\arrayrulewidth}
      \multicolumn{7}{c}{\textbf{Train with Different Replacement Rates}} \\
      \hline
      NLSSum w/ 0.45                    & 33.09           & 21.75           & 24.13           & 9.84           & 32.42           & 24.25  \\
      NLSSum w/ 0.50                    & 33.43           & 21.78           & \textbf{24.17}  & \textbf{9.99}  & 32.31           & 24.34  \\
      NLSSum w/ 0.55                    & \textbf{33.51}  & 21.74           & 24.10           & 9.91           & \textbf{32.58}  & \textbf{24.37}  \\
      NLSSum w/ 0.60                    & 33.50           & \textbf{21.81}  & 23.98           & 9.86           & 32.32           & 24.29  \\
      \Xhline{4\arrayrulewidth}
    \end{tabular}
    }
  \end{center}
  \caption{Ablation Study, Zero-Shot ROUGE-L Results on Validation Dataset of MLSUM}
  \label{result_ablation}
\end{table}

To investigate the influence of each components in NLSSum, we conduct
experiments on the validation set of MLSUM and the results are in Table
\ref{result_ablation}.
In neural label search, we have two weight predictors, the sentence level
predictors \(\mathcal{T}_{\alpha}\) and the label set level predictor
\(\mathcal{T}_{\beta}\) (Section \ref{sec_sent_level}).
We can see from the first block of Table \ref{result_ablation} that without
\(\mathcal{T}_{\beta}\), the result of NLSSum drops.
NLSSum leverages four label sets ($U_a$, $U_b$, $U_c$ and $U_d$) to train
\(\mathcal{T}_{\alpha}\) and \(\mathcal{T}_{\beta}\).
In the second block, we study the effect of each label set separately (note that
XLMRSum-WR is the backbone of NLSSum and we therefore build label set baselines
upon it).
$U_a$ works best overall. However, $U_b$ is better on Russian compared to $U_a$,
which indicates these different label sets can compensate for each other.
Not surprisingly, using one label set performs worse than NLSSum.
In the third block, we use all the label sets, but we use \emph{fixed weights}
instead of using weight predicted from neural label search\footnote{\emph{Fixed
    weight} means a fixed weight for label sets $U_b$, $U_c$ and
  $U_d$, instead of the label search in Section \ref{sec_sent_level}.
  Weight of original English labels $U_a$ is set to 1.0, since the second block
  shows the quality of $U_a$ is the highest.
}.
We can see using multiple label sets can improve variants with only one label
set, but there is still a gap to NLSSum, which learns these weights for each
sentence automatically.
It is also possible to use different weights for different label sets.
To make the number of experiments tractable, we conduct experiments on German
only and search weight around our optimal value (i.e., 0.8).
Results are in Table \ref{grid_search}.
There is slight gain by using different weights, but the result is still worse
than NLSSum.
In the last block, we train NLSSum with different word replacement rates.
We observe that 55\% is the best choice for the bilingual dictionary word
replacement and the word replacement rate is not sensitive.
In practice, we set the rate to 50\% directly instead of tuning it, in order to make the our experiments in  \emph{true} zero-shot settings \cite{perez2021true}.

\begin{table}
  \centering
  \resizebox{0.45\columnwidth}!{
  \begin{tabular}{ccc|rr}
    \Xhline{4\arrayrulewidth}
    \(b\)               & \diagbox{\(c\)}{\(d\)}  & 0.7             & 0.8   \\
    \midrule
    \multirow{2}*{0.7}  & 0.7                     & 32.46           & 32.51 \\
                        & 0.8                     & 32.33           & 32.49 \\
    \midrule
    \multirow{2}*{0.8}  & 0.7                     & 32.94           & \textbf{33.03} \\
                        & 0.8                     & 32.62           & 32.86 \\
    \Xhline{4\arrayrulewidth}
  \end{tabular}
  }
  \caption{ROUGE-L Results for Different Weights}
  \label{grid_search}
\end{table}

\subsection{Human Evaluation}

The human evaluation is important for summarization tasks, since the ROUGE can
only determine the textual representation overlapping.
In this subsection, we design the ranking experiment
\cite{cheng2016neural} with system outputs of different systems on the German
test set of MLSUM.
First, we randomly select 20 samples from the test set of German.
Then, we extract summary sentences from the original document with four mBERTSum,
XLMRSum, NLSSum, and Oracle.
Third, we translate the document and summaries into English by Machine
Translation.
Finally, the human participants are presented with one translated English
document and a list of corresponding translated summaries produced by different
approaches.
Each example is reviewed by five different participants separately.
Participants are requested to rank these summaries by taking the importance and
redundancy into account.
To measure the quality of MT System, we first translate the English document
into German and then back-translate it into English.
We observed that there are almost no changes in meanings between the original
English documents and the back-translated English documents.
We therefore conclude the German to English translation quality is acceptable.
As shown in Table \ref{human_evaluation}, NLSSum is ranked 1st 28\% of the
time and considered best in the extractive models except for Oracle.

\subsection{Monolingual Label Bias}
\begin{figure}
  \centering
  \begin{minipage}[b]{.4\linewidth}
    \includegraphics[width=\linewidth]{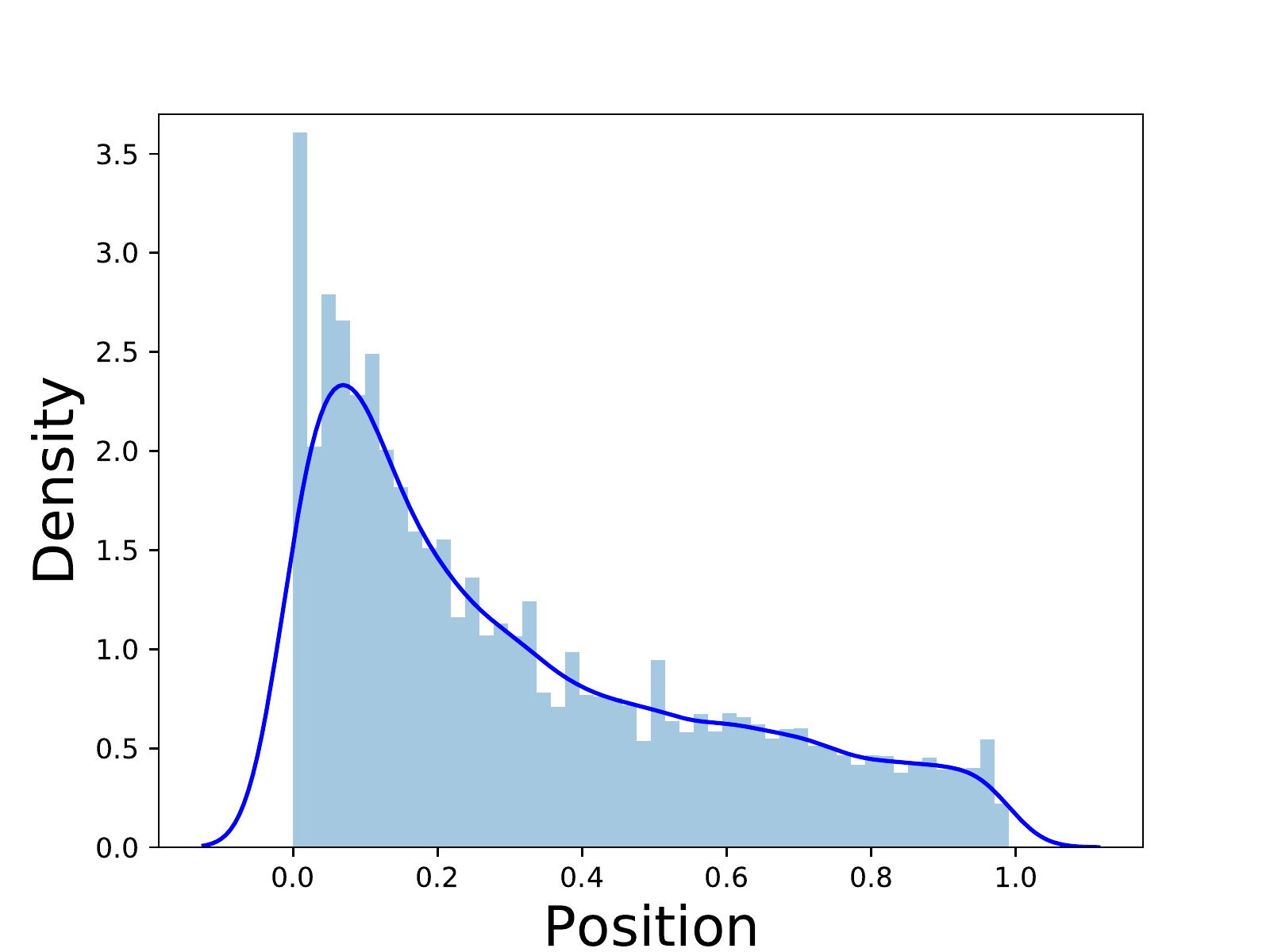}
    \subcaption{English}
    \label{en}
  \end{minipage}
  \begin{minipage}[b]{.4\linewidth}
    \includegraphics[width=\linewidth]{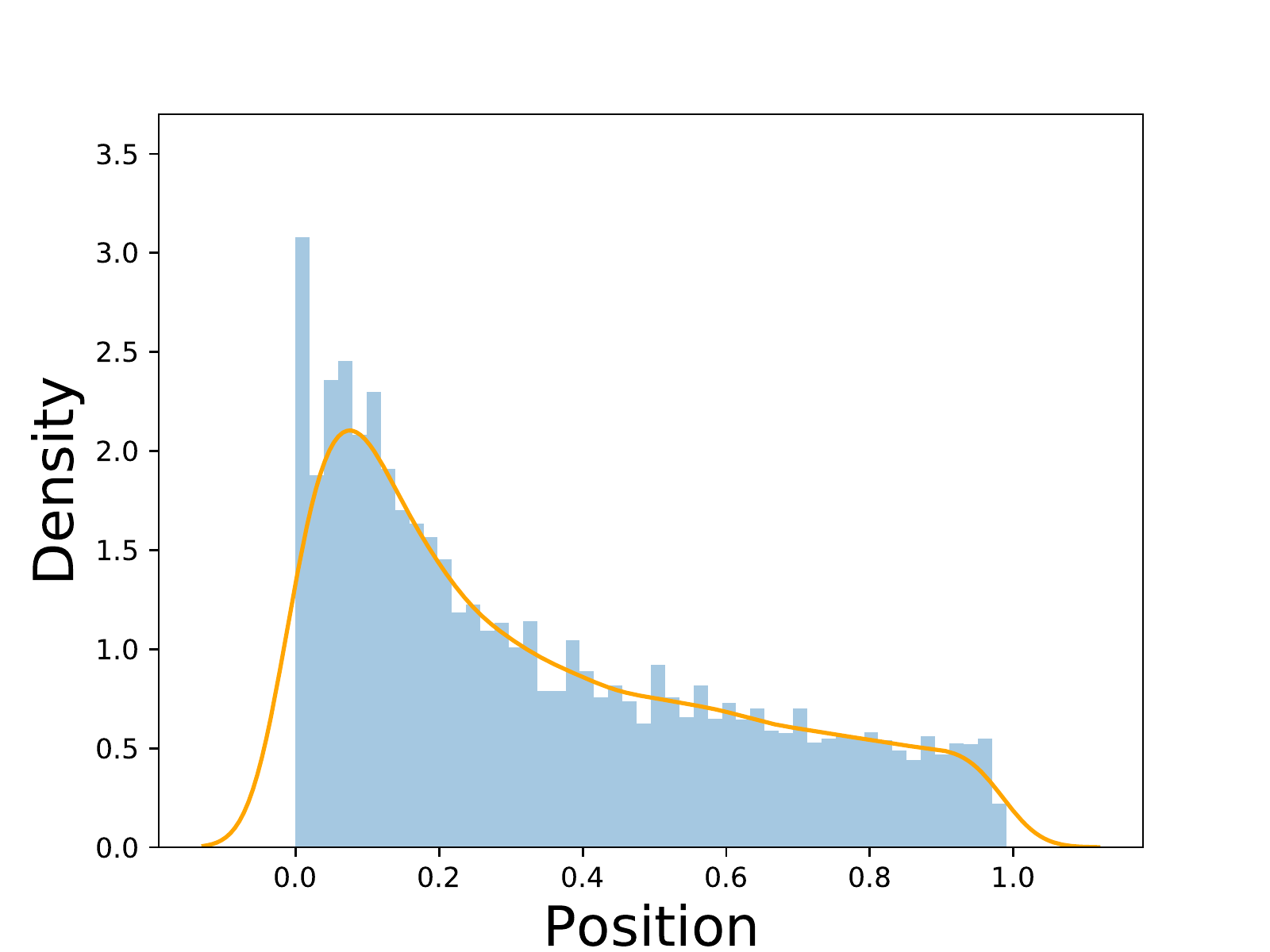}
    \subcaption{Russian}
    \label{ru}
  \end{minipage}
  \begin{minipage}[b]{.4\linewidth}
    \includegraphics[width=\linewidth]{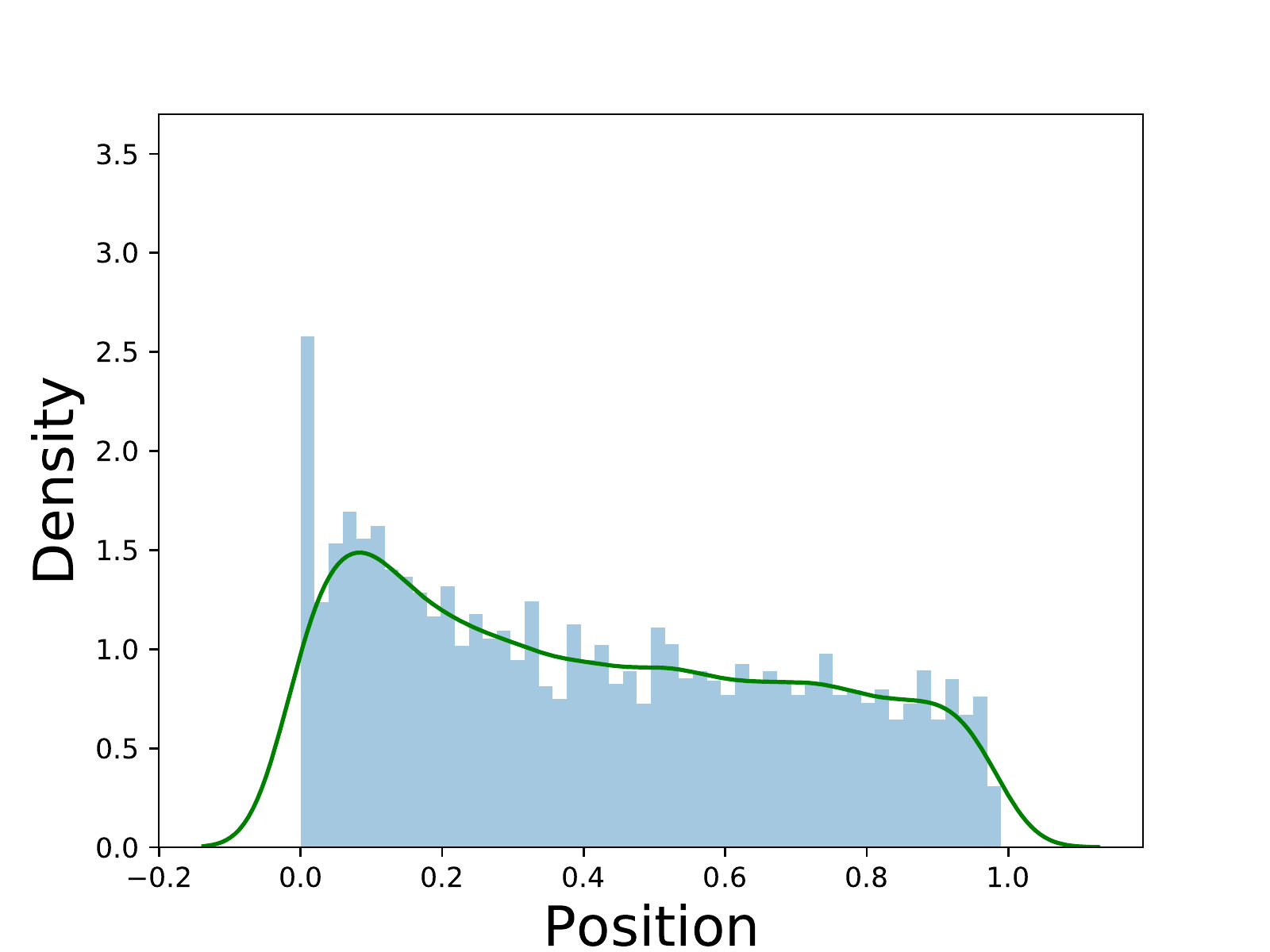}
    \subcaption{Turkish}
    \label{tr}
  \end{minipage}
  \begin{minipage}[b]{.4\linewidth}
    \includegraphics[width=\linewidth]{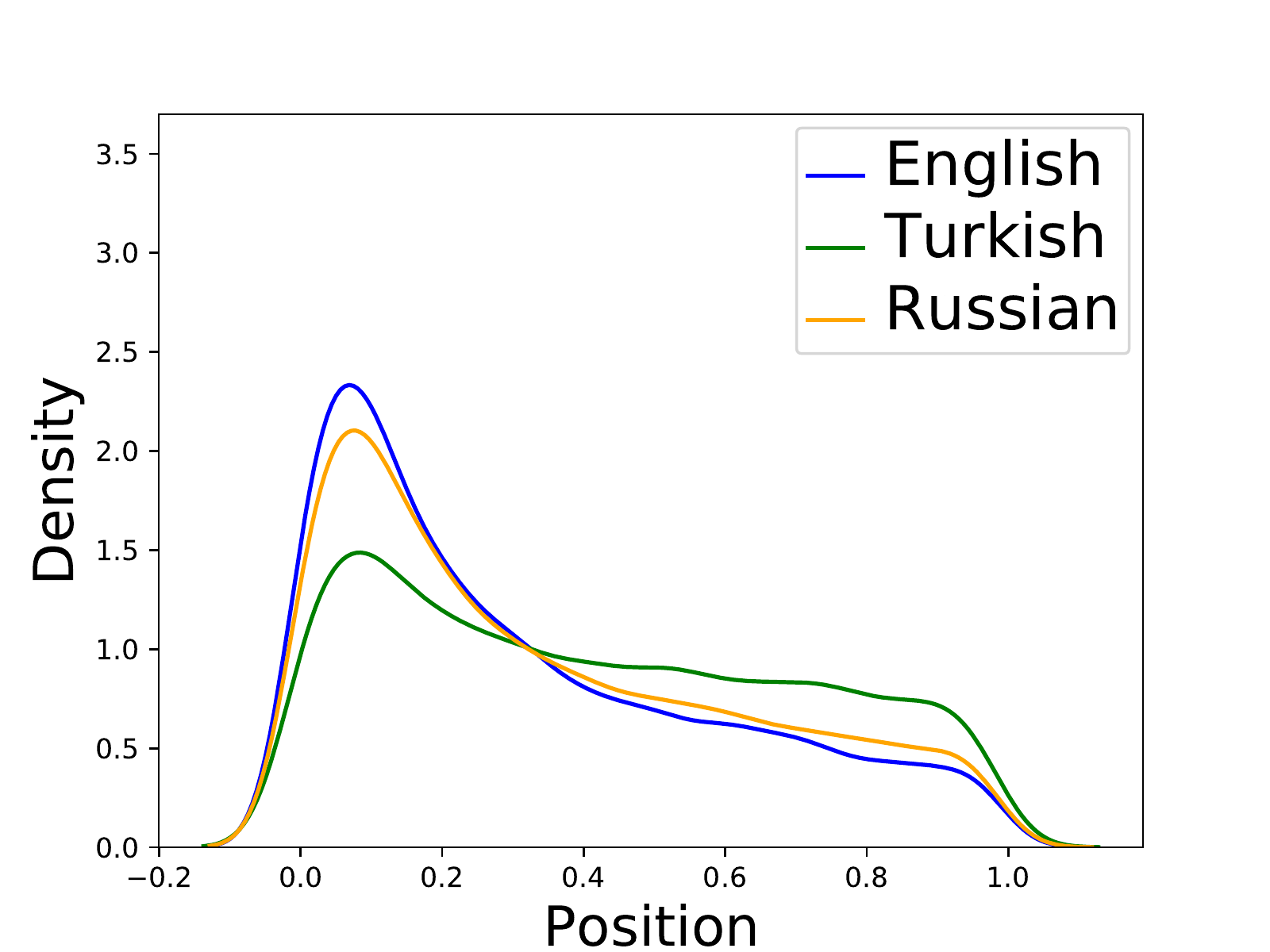}
    \subcaption{Comparison}
    \label{kde}
  \end{minipage}
  \caption{Density of Summary Sentences in CNN/DM}
  \label{visualization}
\end{figure}

In Figure \ref{visualization}, we calculate the positions of oracle sentence
and plot the kernel
density\footnote{https://en.wikipedia.org/wiki/Kernel\_density\_estimation}.
Specically, we translate the test set of CNN/DM from English into Turkish
and Russian, and re-calculate the oracle labels for each language.
Then, we collect all of the oracle sentences and keep its relative positions.
It is obvious that:
1) The oracle sentences of English are mainly located in the head of document,
and the Russian takes the second place, and then the Turkish.
That is why the Turkish achieves more improvement than Russian, by comparing the
results of NLSSum and XLMRSum in the in Part III of Table \ref{result_mlsum}.
2) Multilingual labels pay more attention to the latter sentences, which is more
suitable in multilingual summarization.

\section{Conclusion}
We first study the monolingual label bias, that when translate
the (document, summary) from English into other language, the re-converted
labels will change along with the transformation of textual representation.
Then we propose NLSSum to improve the performance of multilingual zero-shot
extractive summarization, by introducing multilingual labels.
Finally, the summarization model is trained on English with the weighted
multilingual labels and achieves great improvement on other languages.

\section*{Acknowledgements}
This work is supported by the Youth Innovation Promotion Association of the Chinese Academy of Sciences (No. 2018192).

\bibliography{acl2022}
\bibliographystyle{acl_natbib}

\end{document}